\documentclass[lettersize,journal]{IEEEtran}
\usepackage{amsmath,amsfonts}
\usepackage{array}
\usepackage[caption=false,font=normalsize,labelfont=sf,textfont=sf]{subfig}
\usepackage{textcomp}
\usepackage{stfloats}
\usepackage{url}
\usepackage{verbatim}
\usepackage{graphicx}
\usepackage{cite}

\usepackage{todonotes}

\usepackage{xcolor,colortbl}
\definecolor{LightCyan}{rgb}{0.9,0.9,0.9}
\usepackage{arydshln}
\usepackage[linesnumbered,ruled,vlined]{algorithm2e}
\usepackage{graphicx}

\usepackage{enumitem}
\usepackage{algpseudocode}
\usepackage{adjustbox,booktabs,multirow}
\newcommand{\modelname}{\texttt{HateMAML}}

\newcommand{\task}{\mathcal{T}}
\newcommand{\loss}{\mathcal{L}}
\newcommand{\lossi}{\loss_{\task_i}}
\newcommand{\learner}{\texttt{M}}
\newcommand{\data}{\mathcal{D}}
\newcommand{\support}{\mathcal{S}}
\newcommand{\query}{\mathcal{Q}}

\usepackage{amssymb}
\usepackage{amsmath,bm}

\SetKwInput{KwInput}{Input} 
\SetKwInput{KwOutput}{Output} 

\begin{document}

\title{Model-Agnostic Meta-Learning for Multilingual \\Hate Speech Detection}

\author{Md Rabiul Awal, Roy Ka-Wei Lee,  Eshaan Tanwar, Tanmay Garg, Tanmoy Chakraborty 
\thanks{Md Rabiul Awal was with University of Saskatchewan, Canada (e-mail: mda219@usask.ca).}
\thanks{Roy Ka-Wei Lee was with Singapore University of Technology and Design, Singapore (e-mail: roy\_lee@sutd.edu.sg).}
\thanks{Tanmoy Chakraborty was with IIT Delhi, India (email: tanchak@iitd.ac.in)}
\thanks{Tanmay Garg was with IIIT-Delhi, India (e-mail: tanmay17061@gmail.com).}
\thanks{Eshaan	Tanwar was with DTU-Delhi, India (e-mail: eshaantanwar\_2k18co125@dtu.ac.in).}
\IEEEcompsocitemizethanks{\IEEEcompsocthanksitem Tanmoy Chakraborty would like to thank the support of the Ramanujan Fellowship (SB/S2/RJN-073/2017), funded by the Science and Engineering Research Board (SERB), India.}
}

\maketitle
\begin{abstract}
Hate speech in social media is a growing phenomenon, and detecting such toxic content has recently gained significant traction in the research community. Existing studies have explored fine-tuning language models (LMs) to perform hate speech detection, and these solutions have yielded significant performance. However, most of these studies are limited to detecting hate speech only in English, neglecting the bulk of hateful content that is generated in other languages, particularly in low-resource languages. Developing a classifier that captures hate speech and nuances in a low-resource language with limited data is extremely challenging. To fill the research gap, we propose \modelname, a model-agnostic meta-learning-based framework that effectively performs hate speech detection in low-resource languages. \modelname\ utilizes a self-supervision strategy to overcome the limitation of data scarcity and produces better LM initialization for fast adaptation to an unseen target language (i.e., cross-lingual transfer) or other hate speech datasets (i.e., domain generalization). Extensive experiments are conducted on five datasets across eight different low-resource languages. The results show that \modelname\ outperforms the state-of-the-art baselines by more than 3\% in the cross-domain multilingual transfer setting. We also conduct ablation studies to analyze the characteristics of \modelname. 
\end{abstract}
\begin{IEEEkeywords}
Hate speech detection, cross-lingual transfer, meta-learning.
\end{IEEEkeywords}

\section{Introduction}

\IEEEPARstart{M}{otivation.} Online hate speech that expresses hate or encourages violence towards a person or a group based on characteristics such as race, religion, or gender is a growing global concern. The hateful content has broken the cohesiveness of online social communities and resulted in violent hate crimes\footnote{https://thewire.in/law/hate-speech-what-it-is-and-why-it-matters}. Therefore, detecting and moderating hate speech in online social media is a pressing issue. 

The gravity of the situation has motivated social media platforms and academic researchers to propose traditional machine learning and deep learning solutions to detect online hate speech automatically \cite{park-fung-2017-one,10.1145/3394231.3397890,awal2021angrybert} and early~\cite{lin2021early,meng2022predicting}. Among the solutions, large pre-trained language models (LMs), which have demonstrated their superiority in many NLP tasks \cite{devlin2019bert}, have also shown excellent performance in the hate speech detection task \cite{ranasinghe-zampieri-2020-multilingual}. However, existing studies have predominantly focused on detecting hate speech in English and content from Western cultures. This neglects the large portion of online hate speech from other languages and cultures. Low-resource languages often have very limited or no training samples available. This can make it challenging to develop supervised classifiers. We need models that can efficiently adapt with few training data or methods that can transfer knowledge from another dataset. In such a case, a high-resource language can augment the low-resource language by knowledge transfer. For example, Aluru et al. \cite{Aluru2020ADD} proposed a method to fine-tune a multilingual BERT model for hate speech detection in low-resource languages that leverages cross-lingual transfer from high-resource pre-training.


There are very few works that deal with multilingual hate speech detection. A viable approach is to fine-tune pre-trained LMs, which is explored in existing studies~\cite{ranasinghe-zampieri-2020-multilingual, pamungkas-patti-2019-cross, Aluru2020ADD}. The underlying intuition is that the large LMs generate shared embeddings in many languages,  enabling cross-lingual transfer from supervised training in the {\it high-resource} languages such as English. This cross-lingual transfer can be achieved as follows: (i) training only in the source language and evaluating in target language, known as `zero-shot', or (ii) training in the target language only or joint training of source and target languages, known as `few-shot' learning. In multilingual hate speech detection, English and a few high-resource languages are considered ``sources'', and low-resource languages of interest are considered ``targets''. A source language is usually a magnitude larger than a target language. However, the fine-tuning of pre-trained LMs also has several limitations~\cite{nozza-2021-exposing}. An obvious issue is the data scarcity in target low-resource languages; it is challenging to capture hate speech and its nuances with insufficient observations. The performance of cross-lingual transfer learning may adversely be affected if source and target languages are from distant language families \cite{annotatorbiaskuwatly,nozza-2021-exposing}. 


Hate speech datasets can come from many domains, which can be divided into two levels: one comes from language shift, and the other is content in each language. For instance, a dataset in one language may be collected from the Twitter platform, while another could be collected from Youtube. Standard fine-tuning overfits the training languages and will face the difficulty of adapting to a new domain or an unseen language. Conneau et al. \cite{conneau-etal-2020-unsupervised} show that adding more languages improves the performance on low-resource languages but hurts the performance on high-resource languages. We adopt a meta-training strategy for low-resource hate speech detection to address these concerns. Our proposed model is based on a meta-learning approach that can be used for low-resource hate speech detection. The model is capable of handling domain adaptation and preventing negative transfer. Meta-learning has been successful in low-resource settings and can be used to learn a good initial point for fine-tuning a new task, making the fine-tuning process more efficient.


\textbf{Research objectives.} In this paper, we aim to tackle the challenges of multilingual hate speech detection and address the limitations of existing fine-tuned pre-trained LMs in hate speech detection. We propose $\modelname$, a model-agnostic meta-learning-based framework that effectively performs hate speech detection in low-resource languages. Unlike existing hate speech detection approaches that fine-tune LMs on the source and target only, we focus on resource maximization by using resources in other ``\textit{auxiliary}'' languages beyond source and target. More specifically, we adopt the model-agnostic meta-learning (MAML) \cite{10.5555/3305381.3305498} strategy, in which simulated training in few-shot meta-tasks produces a meta-learner that can learn a target task quickly.

Recent studies have explored adopting MAML in low-resource cross-lingual adaptation for NLP tasks~\cite{gu-etal-2018-meta, nooralahzadeh-etal-2020-zero}. The underlying intuition of $\modelname$ is to train on samples from auxiliary languages and adapt it to detect hate speech in a target language in which there are few or no datasets available \cite{nooralahzadeh-etal-2020-zero}. To address diversity and nuances in detection of hate speech across languages and domains, we also modify the meta-learning loss function in $\modelname$ and add a domain adaptation loss. Intuitively, task-specific learner evaluates their own training samples and samples from another language. This mimics the scenario we are interested in: training in one language while evaluating an unseen language during test time. Such training objective forces the meta-learner to generalize to a target language domain while learning from source domain samples. 


Considering the data scarcity of non-English languages, we further propose a self-guided meta-learning-based fine-tuning mechanism. This mechanism utilizes unlabeled examples of the low-resource target language and generates silver labels for fine-tuning the target. We first train a multilingual predictor model using training data in a high-resource language, e.g., English. The predictor model predicts unlabeled samples in the target, e.g., Spanish. A small portion of quality predictions are filtered out using a threshold value to gather silver labels and enhance the training data in both the source and target languages. Finally, we adapt $\modelname$ into an iterative self-refining loop that replaces and refines the current predictor model and improves the quality of the adapted model in the target.

In early studies, cross-lingual meta-training was mostly limited to language pairs \cite{nooralahzadeh-etal-2020-zero}. This setting is, however, not truly multilingual since we need separate models for each target language. We study the applicability of meta-training when training data is available in more than one low-resource language. To this end, we investigate whether meta-training is better than standard multilingual fine-tuning in two scenarios: when training examples are only available for some languages and when both high and low-resource languages have sufficient training data. The first scenario is a stress test for cross-lingual domain adaptation, evaluating the trained model on a set of reserved low-resource languages. The other scenario explores meta-training benefits over standard fine-tuning. Though the early investigation was limited to one auxiliary and target language, the current configuration allows us to investigate cross-lingual meta-training for hate speech detection at scale. In other words, we adopt meta-training as a proxy for general multilingual fine-tuning. Our results show that when training data is available from many languages, meta-training $\modelname$  yields better performance than standard fine-tuning.


\textbf{Contributions.} We present $\modelname$, a meta-learning framework for multilingual hate speech detection with limited data at scale. Our major contributions are summarized below: 

\begin{enumerate}[leftmargin=*]
    \item We propose $\modelname$\footnote{https://gitlab.com/bottle\_shop/safe/HateMAML}, a domain-adaptive MAML algorithm for hate speech detection in low-resource languages. Our algorithm efficiently exploits the available resources, outperforming state-of-the-art cross-lingual fine-tuning baselines. One of the novelties of $\modelname$ lies in the efficient utilization of available samples in the validation set of source and training data in auxiliary languages. 
    \item To mimic an extreme data-scarce scenario in the target language (zero-shot), we introduce a self-refining variant of $\modelname$. We adopt an iterative meta-learning approach to generate silver labels from unlabeled data in the target language.
    \item We conduct extensive experiments in zero-shot and fine-tuning settings across eight different low-resource languages. Our experiment results show that $\modelname$ consistently outperforms state-of-the-art methods by more than 3\% in the cross-lingual transfer settings. 
    \item We perform ablation studies, in which we carefully examine the contribution of the auxiliary languages across each \textit{auxiliary- target} language pair. We also assess the algorithm's robustness by varying (i) the amount of meta-training samples, (ii) cross-domain adaptation, and (iii) one model to support all language configurations. 
    \item We conduct cross-domain hate speech detection to address data set diversity in various languages and show that \modelname\ achieves a significant domain adaptation. Our experiments on meta-training over all available language training data suggest that meta-learning could be an alternative to standard fine-tuning that yields superior performance.
\end{enumerate}


\section{Related Work}
{\bf Multilingual hate speech detection} 
In the last few years, hate speech detection has gained significant attention, with the majority of work dedicated towards monolingual hate speech \cite{ park-fung-2017-one, 10.1145/3394231.3397890}. This skew in attention has naturally led to the majority of datasets being in English \cite{DBLP:conf/icwsm/DavidsonWMW17, waseem-hovy-2016-hateful, founta2018large}. However, hate speech is a global phenomenon, and increased diversity in available resources is critical for developing automated systems. Recent efforts   addressed this issue, including multiple shared tasks (such as SemEval 2020 \cite{zampieri-etal-2020-semeval}, HASOC 2020 \cite{10.1145/3368567.3368584}, Evalita 2018 \cite{Fersini2018OverviewOT} and HateEval 2019 \cite{basile-etal-2019-semeval}). Such efforts helped in bringing much-needed traction toward multilingual hate speech research. Additionally, recent development in transformer-based large multilingual LMs (mBERT \cite{devlin2019bert}, XLM-R \cite{conneau-etal-2020-unsupervised}), pre-trained on more than 100 languages, has aided in developing state-of-the-art classifier in resource-lean languages. Prior studies on multilingual hate speech detection have explored: (i) resource development such as creating datasets \cite{zampieri-etal-2020-semeval,10.1145/3441501.3441517}, shared tasks and organizing workshops, (ii) cross-lingual transfer learning with multilingual shared embeddings and pre-trained LMs \cite{pamungkas-patti-2019-cross,Aluru2020ADD}, (iii) utilization of additional features from relevant domains \cite{markov-etal-2021-exploring} (e.g., emotion, sentiment), and (iv) data augmentation \cite{wullach2021fight,bigoulaeva-etal-2021-cross,pamungkas-patti-2019-cross} techniques including use of external service (e.g., translation APIs) for supervised training.

{\bf Cross-lingual and multilingual transfer.}
Knowledge transfer from a high-resource language (i.e., English) to a low-resource one (i.e., Hindi) has been shown to be an effective resource optimization technique. Ranasinghe and Zampieri \cite{ranasinghe-zampieri-2020-multilingual} used a two-step training method: first training multilingual transformer models on English data, and then the resulting model is fine-tuned on the low resource target language. Such multi-phase training was found to be a better model initialization for fine-tuning the target. Aluru et al. \cite{Aluru2020ADD} conducted an extensive study on state-of-the-art multilingual embeddings including LASER \cite{10.1162/tacl_a_00288} and MUSE \cite{conneau2017word}. Another line of studies  \cite{pamungkas-patti-2019-cross,10.1145/3465336.3475102,8955559} utilized translation-based solutions to alleviate the shortage of data in low-resource languages in hate speech. These models rely heavily on translation APIs, which might be susceptible to producing low-quality translations. Moreover, the translation of a large source corpus can be expensive.

Majority of the studies on cross-lingual and multilingual hate detection fall into pre-trained model fine-tuning \cite{DBLP:journals/corr/abs-1909-12642,Mishra20193IdiotsAH,10.1145/3465336.3475102}. However, Nozza \cite{nozza-2021-exposing} found that due to the domain shift of hate speech across languages, standard fine-tuning methods are less effective for both zero-shot and few-shot settings. 
To address the limitation of low-resource language fine-tuning, {\em we develop $\modelname$ for fine-tuning hate classifiers with limited data and an improved mechanism for domain-adaptive cross-lingual transfer.}

{\bf Meta-learning.}
Meta-learning, also known as ``learning to learn'', is a few-shot learning technique geared toward learning few-shot tasks and fast adaptation. Unlike standard supervised fine-tuning on downstream tasks that result in a \textit{final} classifier, meta-learning focuses on producing a rich initialization point from where an unseen target task can be learned quickly. Some common meta-learning approaches  include (i) optimization-based \cite{10.5555/3305381.3305498,gordon2019metalearning}, and (ii) metric-based \cite{Koch2015SiameseNN,NIPS2016_90e13578}. Finn et al.  \cite{10.5555/3305381.3305498} introduced model-agnostic meta-learning (MAML), an optimization-based meta-learning framework for few-shot tasks. Several recent studies have explored gradient-based meta-learning for few-shot text classification \cite{yu-etal-2018-diverse,han-etal-2021-meta,bao2020fewshot}. Meta-learning for domain generalization is also studied to handle domain shift during training in diverse problems, including supervised learning and reinforcement learning \cite{Li2018MLDG}. Meta-learning has also been explored in zero-shot and few-shot cross-lingual settings. Gu et al. successfully adapted MAML for low-resource NMT tasks using auxiliary languages, resulting in competitive performance with only a fraction of the training data \cite{gu-etal-2018-meta}. X-MAML was then proposed by \cite{nooralahzadeh-etal-2020-zero} for meta-training in cross-lingual NLU tasks, similar to our interest. X-MAML explores various auxiliary languages to identify the optimal composition for zero-shot cross-lingual transfer. Meta-learning has also been applied in the detection of offensive language in cross-lingual and code-mixed texts \cite{9696324,10.1145/3503162.3503167} and other harmful content such as multilingual rumours~\cite{awal2022muscat}. However, the limited availability of multilingual hate speech datasets, comprising of only two or three languages, presents a challenge in finding an effective auxiliary language.


{\bf Data augmentation/self-training.}
Data augmentation approaches can be divided into three categories: (i) rule-based \cite{wei-zou-2019-eda}, (ii) interpolation-based \cite{wang-etal-2018-switchout,guo-etal-2020-sequence,jindal-etal-2020-augmenting}, and (iii) model-based \cite{sennrich-etal-2016-improving,feng-etal-2019-keep,DBLP:conf/naacl/Kobayashi18}. In cross-lingual hate speech, translation techniques \cite{pamungkas-patti-2019-cross,10.1145/3465336.3475102,8955559} are commonly used for data augmentation during training. Rather than collecting and annotating new hate speech data, bootstrapping on
unlabeled samples to create pseudo-labeled data provides a way for data augmentation and fine-tuning on low-resource languages. We draw inspiration from \cite{bigoulaeva-etal-2021-cross} to device a bootstrapping-based self-training approach for $\modelname$.

{\bf How is our approach different?}
We explore cross-lingual meta-training in the domain of hate speech for both zero-shot and fine-tuning configurations. Our proposed method of zero-shot meta-training, \modelname, is a novel idea that can be further adapted for languages with no availability of labeled data. We focus on resource maximization and domain generalization while transferring task-specific knowledge to low-resource languages. We carry out a large-scale study in multilingual hate speech detection across diverse domains on available hate speech datasets. 


\section{Methodology}

We devise a gradient-based meta-learning algorithm named $\modelname$ for multilingual hate speech detection. Our proposed algorithm is tailored to improve cross-lingual transfer in target low-resource languages. We create a set of meta-tasks from samples for both high- and low-resource languages and simulate episodic meta-training similar to MAML \cite{10.5555/3305381.3305498}. Our goal is to produce better initialization of model parameters that can adapt to a low-resource target language task using -- (i) none (i.e., zero-shot) or (ii) only a small number (i.e., few-shot) of labeled examples. We select an auxiliary language to (meta-)train the model in the zero-shot setting.
In the few-shot setting, we meta-train the model on the target language. We assume some labeled data (e.g., 2024 samples) is available for supervised fine-tuning on the target language. 
Hate speech datasets from different languages often comprise several domains and topics.
For example, while many English datasets capture the domain of racism-related hate, hate speech datasets in Hindi are more religion/politics-oriented.
Here, the term `domain' is loosely used, referring to different datasets and languages. Therefore, the task of cross-lingual hate speech detection implicitly encompasses cross-domain adaptation. We aim to {\em fast adapt to both target language and domains}. To account for this, our meta-adaptation model includes a domain generalization loss. The idea is to produce a good initialization that can perform well on diverse domains.

\subsection{Model Agnostic Meta-Learning (MAML)}
\label{sec:maml}
MAML assumes a distribution $p(\mathcal{T})$ of tasks $\{\mathcal{T}_1, \mathcal{T}_2, \cdots, \mathcal{T}_m\}$. A meta-model $f_\theta(\cdot)$ with parameters $\theta$ is learned through episodic meta-training over sample tasks $\mathcal{T} \sim p(\mathcal{T})$.  MAML has two loops -- (i) an inner loop for task-specific adaptation and (ii) an outer loop to \textit{meta-learn} that fast adapt to an unseen novel task.

MAML fine-tunes the meta model $f_\theta$ to a particular task $\mathcal{T}_i$ through gradient descent:
\begin{equation}
    \theta_i^\prime \leftarrow \theta - \alpha \nabla \mathcal{L}_{\mathcal{T}_i}(f_\theta)
    \label{eq:1}
\end{equation}
where $\alpha$ is the step size, and $\mathcal{L}$ is the classification loss. The task-specific training outcome is evaluated on the associated test set. 

The meta-learner optimization objective is to minimize the \textit{meta loss} computed from all the training tasks:

\begin{equation}
    \min_{\theta} \sum_{i=1}^m \mathcal{L}_{\mathcal{T}_i} (f_{\theta^\prime}) = \sum_{i=1}^m \mathcal{L}_{\mathcal{T}_i} (f_{\theta - \alpha \nabla_\theta \mathcal{L}_{\mathcal{T}_i}(f_\theta)}).
    \label{eq:2}
\end{equation}

The meta parameter $\theta$ is then updated to:
\begin{equation}
    \theta = \theta - \beta \nabla_\theta \sum_{i=1}^m \mathcal{L}_{\mathcal{T}_i} (f_{\theta_i^\prime})
    \label{eq:3}
\end{equation}
where $\beta$ is the meta-learner learning rate. We can accumulate multiple episodes of tasks to update $\theta$. Though MAML training requires double gradient descent optimization, a first-order approximation is used in practice. Different from standard fine-tuning, meta-training does not result in a final model. However, it is a reasonably good starting point (initialization) from which learning a new task can be executed quickly. Intuitively, training across a set of meta-tasks can be seen as \textbf{auxiliary}; and fast adaptation on the unseen \textbf{target} task is the main goal. A few recent studies build on this motivation \cite{nooralahzadeh-etal-2020-zero, gu-etal-2018-meta} where supports from auxiliary languages are used in cross-lingual meta-training. This is also a key inspiration for our model.

\begin{figure*}
    \centering
    \includegraphics[width=0.875\linewidth]{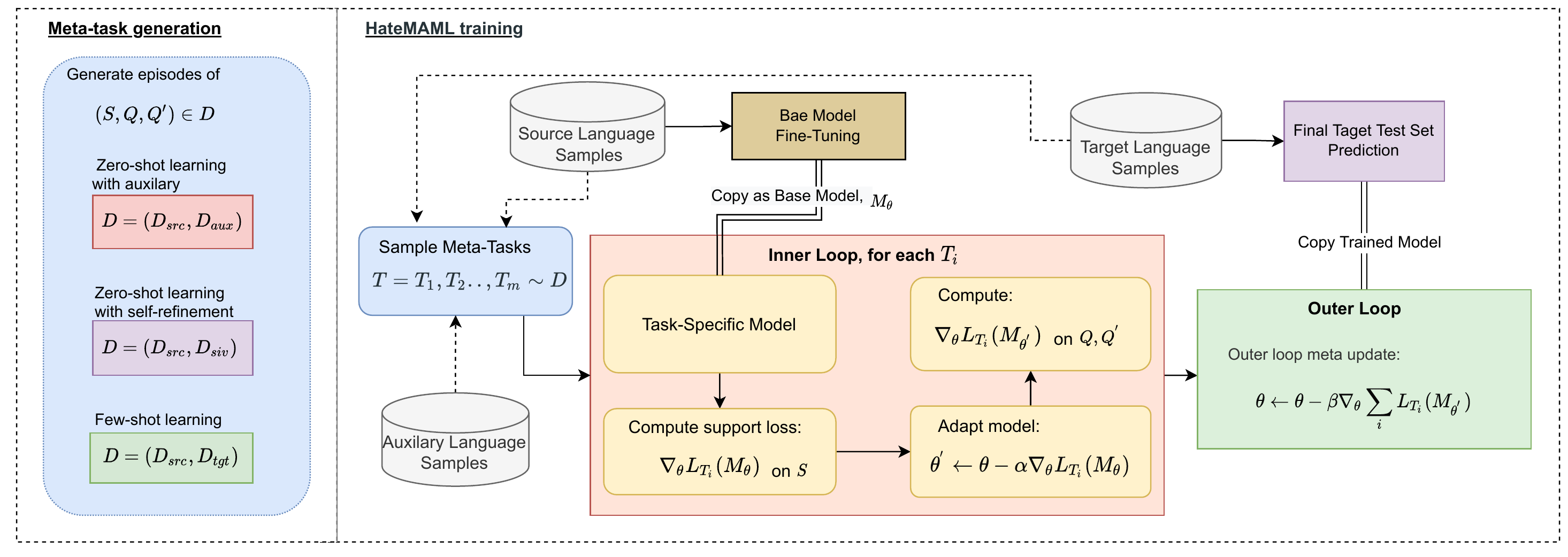}
    \vspace{-5mm}
    \caption{Proposed $\modelname$ model sketch. (Left) Model training in zero-shot and few-shot setups. We use an auxiliary language validation set in zero-shot learning, and target language labeled data is used in few-shot. In both cases, we include a source language validation set and create a (virtual) domain set that helps us to mimic target domain performance evaluation during meta-training. We sample a batch of triplets $(\support, \query, \query^\prime)$ and simulate episodic training. (Right) Self-training procedure with $\modelname$, which produces silver labels from unlabeled data in the target language, model training on those, replace base model. Repeat the procedure for $N$ times.}
    \label{fig:model_sketch}
\end{figure*}

\subsection{Proposed Model: \modelname}
\label{sec:hatemaml}
The training of our proposed \modelname\ model facilitates hate classifier adaptation and cross-lingual transfer in target languages while requiring only a small number of samples for model fine-tuning. \modelname\ can be used for training with/without labeled data, only requiring access to training data in the source language for the initial fine-tuning. Any multilingual encoders that have shared embeddings, such as mBERT and XLM-R can be used as the base model. To further improve cross-lingual transfer, we use a few labeled samples in the target language (few-shot training), which helps quick adaptation. Additionally, \modelname\ benefits from task-specific training evaluation on a (virtual) domain test (query) set at each episode that mimics target domain evaluation during meta-training \cite{Li2018MLDG}. The intuition of the domain query set is to achieve training generalization in unseen target languages and domains. We discuss training datasets accumulation, meta-tasks generation, and domain loss in $\modelname$ and present a concise training procedure in detail below. A rough sketch of $\modelname$ can be found in Figure ~\ref{fig:model_sketch}(right).

{\bf $\modelname$ Training Data.}
We need a set of support and query batches sampled from $\data$, where $\data$ refers to available training data from training languages, generally small in size. Suppose we have samples in  source  $(\data_{src})$, auxiliary $(\data_{aux})$ and target $(\data_{tgt})$ languages. Our training data $\data$ consists of the tuple: (i) $(\data_{src}, \data_{aux})$ in the zero-shot setting, and (ii) $(\data_{src}, \data_{aux})$ in the few-shot setting. To this end, we split the training set for each language into a support ($\data_{lang}^{support}$) and query ($\data_{lang}^{query}$), where $lang \in \{src, aux, tgt\}$. We randomly sample a (virtual) domain set $\data^{domain-query}$ from either $\data_{aux}^{query}$ or  $\data_{tgt}^{query}$. By doing so, we can mimic real train-test domain shifts so that over many iterations, we can train a model to achieve better generalization on the target language's \textbf{final} test set. 

{\bf Meta-Tasks Generation.}
\modelname\ requires episodic training on meta-tasks containing support (meta-training) and query (meta-testing) sets. For each episode, three subsets of examples are sampled separately: the support set $\support$, the query set $\query$, and the (virtual) domain query set $\query^\prime$.
\begin{equation}
    \begin{aligned}
        &S \leftarrow S \cup \wedge (X, Y) \in \data \\
        &Q \leftarrow Q \cup \wedge (X, Y)  \smallsetminus S  \in \data \\
        &Q^\prime \leftarrow Q^\prime \cup \wedge (X, Y)  \smallsetminus (S \cup Q) \in \data \\
    \end{aligned}
\end{equation}
We repeat this process multiple times. 
Suppose we have a total of $D$ training samples in a given language and $K$ samples used for support, and $L$ for both query sets in each episode. In total, we will have $D/K+L$ episodes. Note that $\query^\prime$ is virtual and selected randomly.

{\bf (Virtual) Domain Query Loss.}
We add query loss term in Equation ~\ref{eq:3} that accounts for domain generalization. 
We now compute meta-update in Equation \ref{eq:3} as follows:
\begin{equation}
    \theta = \theta - \beta \nabla_\theta \sum_{i=1}^m \mathcal{L}_{\mathcal{T}_i}^{(\mathcal{Q}_i, \mathcal{Q}_i^\prime)} (f_{\theta_i^\prime})
\end{equation}
It is a variation of standard MAML training that addresses supervised training limitations such as domain shift in low-resource languages. We compare our proposed training procedure with MAML to show the effectiveness.

\subsection{HateMAML Algorithm}
\label{sec:algorithm}
$\modelname$ training requires a base classifier. A base classifier can be any multilingual encoder, such as mBERT, XLM-R, fine-tuned on a source language such as English. Next, we decide between zero-shot or few-shot training based on the availability of training data. The training choice is passed to Algorithm ~\ref{alg:HateMAML} along with the base model parameters. For the zero-shot scenario, we select an auxiliary language for model training, and the resulting meta-model will be evaluated on the selected target language. Meta-learner parameters $\theta$ are initialized from a fine-tuned base model (\textit{line 2}). Now, we sample a batch of pseudo-tasks $\mathcal{T}_i$. Each task contains a triplet $(\support, \query, \query\prime)$ representing the support, query, and domain query for model training, respectively. We take one inner loop gradient step using training loss on $\support$ (\textit{line 10}) and adapt model parameters to $\theta^\prime$. Now, we evaluate task-specific training outcomes using $\query$ and $\query^\prime$, and save it for outer loop update ({\it line 12}). Each task has its own task-adaptive model parameters $\theta_i^\prime$. At the end of iterations of all tasks, we aggregate saved gradients and update the meta-model $\theta$ (\textit{line 13}). At this step, we have completed one outer loop. We repeat {\em lines 8-13} till model training is not finished. After training completion, based on the earlier choice of zero-shot or few-shot, we evaluate the resulting model's performance on the target language's {\bf final} test set.

\begin{algorithm}[t]\small
\DontPrintSemicolon
 \KwInput{Training choice \texttt{C}, high-resource language \texttt{h}, set of low-resource languages \texttt{L}, Meta-learner $\learner$,  fast lr $\alpha$, meta lr $\beta$}
  
 Fine-tune $\theta$ on \texttt{h} (base model) \\
 \textbf{Initialize} $\theta \leftarrow  \theta $\\
  \uIf{\texttt{C} is zero-shot}{
    Select auxiliary from \texttt{L} to get $\data = (\data_{src}, \data_{aux})$
  }
  \Else{
    Select target language get \texttt{L} to form $\data = (\data_{src}, \data_{tgt})$
  }
  \While{not done}
    { 
    Sample batch of tasks $\task = \{\task_1,\task_2..,\task_m\} \sim \mathcal{D} $ \\ 
    \For {all $\task_i = (\support, \query, \query^\prime) $ in $\task$} 
    {
    Compute $\nabla_{\theta}\lossi(\learner_{\theta})$ on $S$ \\
    Compute adapted parameters with gradient descent:
    $\theta^{'}= \theta - \alpha\nabla_{\theta}\lossi(\learner_{\theta})$ \\
    Compute $\lossi(\learner_{\theta^{'}})$ using $\query$, $\query_d$ for outer loop meta-update \\
    }
    Update $ \theta \leftarrow \theta - \beta \nabla_{\theta} \sum_{i} \lossi(\learner_{\theta^{'}})$ \\
  }

    Perform test set evaluation on the target language using $\learner_\theta$
 
\caption{\modelname}
\label{alg:HateMAML}
\end{algorithm}
  
\subsection{Self-refinement and Zero-shot Learning}
\label{sec:selftrain}
We now present a semi-supervised self-training procedure shown in Figure \ref{fig:model_sketch} (left) for hate speech classifier adaptation while no target or auxiliary language is available. Our proposed approach requires the availability of the source language. We first develop a `base model' by fine-tuning source language training data. Now, we can use the trained model to make predictions on unlabeled samples from the target language. We want to use these predicted samples for supervised training. Directly including all the predicted samples in training may not result in a good classifier, as training samples are noisy and do not represent ground-truth labels. We filter out samples with low prediction confidence using a threshold value to keep it balanced. We call the filtered outcomes \textit{silver labels} and prepare a new training dataset. We use $\modelname$ to carry out meta-training since it requires a few samples and can adapt quickly. As we use a small number of samples, noise injection is less than standard fine-tuning. We randomly take filtered samples as silver labels and run $\modelname$ using few-shot training shown in Algorithm ~\ref{alg:HateMAML}. After completing one full training on the silver training data, the new classifier should expectantly have better target language knowledge. To make it work in practice requires self-refinement with multiple iterations. We repeat this procedure by treating the resulting model as the base for generating silver labels again, replacing the old base model. In our experiments, approximately $5$ refinement iterations with $\modelname$ are enough for producing a quality classifier on the target test data. 

\begin{table*}[t!]
    \caption{A summary of the multilingual hate speech datasets.}
    \vspace{-3mm}
    \centering
    \adjustbox{max width=1\linewidth}{

    \begin{tabular}{lp{4cm}p{3cm}p{3cm}p{4cm}}
    \toprule
    \textbf{ Dataset } & \textbf{ Language } & \textbf{ Size of training set } & \textbf{Size of test set} &  \textbf{ Domain }\\
    \hline
    Founta-EN\cite{founta2018large}  & English (EN) & 80000 & --- & Twitter stream \\
    \hline
    HatEval19 \cite{basile-etal-2019-semeval} & English (EN), Spanish (ES) & \textbf{EN}: 9000, \textbf{ES}: 4500 & \textbf{EN}: 2971, \textbf{ES}: 1600 & Immigrant, Woman \\
    \hline
    SemEval20 \cite{zampieri-etal-2020-semeval}          & Arabic (AR), Danish (DA), Greek (EL), Turkish (TR)  & \textbf{AR}: 8000, \textbf{DA}: 2961, \textbf{EL}: 8743, \textbf{TR}: 31756 & \textbf{AR}: 200, \textbf{DA}: 329, \textbf{EL}: 1544, \textbf{TR}: 3528  &  Twitter, Facebook, Reddit, News   \\
    \hline
    HASOC20 \cite{10.1145/3441501.3441517} & English (EN), Hindi (HI), German (DE) & \textbf{EN}: 3708, \textbf{HI}: 2963, \textbf{DE}: 2373  & \textbf{EN}: 1628, \textbf{HI}: 1326, \textbf{DE}: 1052 & Twitter stream \\
    \hline
    HaSpeeDe \cite{sanguinetti2020haspeede} & Italian (IT) & \textbf{IT}: 6839 & $\bf{IT_{tweeter}}$: 1263, $\bf{IT_{news}}$: 500 & Immigration, Roma, Muslims \\

    \bottomrule
    \end{tabular}
    }
    \label{tab:dataset_summary}
    \vspace{-5mm}
\end{table*}

\section{Hate speech detection datasets}
\label{sec:dataset}

We train and evaluate the baselines and $\modelname$ using five publicly-available multilingual hate speech datasets. These datasets contain a variety of low-resource languages. Table ~\ref{tab:dataset_summary} summarizes the datasets.

\begin{itemize}[leftmargin=*]
    \item \textbf{Founta-EN} \cite{founta2018large} is a large English dataset of about 80,000 tweets, collected through crowd-sourcing. These tweets were labeled into -- {\it normal}, {\it spam}, {\it abusive}, and {\it hateful}. For our study, we removed the tweets labeled {\it spam}.

\item \textbf{HatEval19} \cite{basile-etal-2019-semeval}, released in the shared task of SemEval-2019, is a multilingual hate speech dataset in English and Spanish, specifically targeted hate against immigrants and women. 

\item \textbf{HASOC20} \cite{10.1145/3441501.3441517} is a multilingual hate speech corpus of three languages -- English, German, and Hindi. The authors first created an extensive archive of tweets and subsequently used weakly trained binary classifiers to extract potential hateful tweets. The extracted ones are finally labeled by annotators. The main aim of the dataset was to create an unbiased multilingual hate speech corpus.

\item \textbf{HaSpeedDe20} \cite{sanguinetti2020haspeede} is a monolingual corpus in Italian. It focuses on hate speech against immigrants, Roma and Muslims. The dataset is collected from two platforms -- Twitter and news headlines. It consists of test sets from Twitter and news.  

\item \textbf{SemEval20} \cite{zampieri-etal-2020-semeval} covers five languages -- English, Arabic, Danish, Greek, and Turkish. Native speakers annotated the dataset. Arabic, Greek, and Turkish were collected from Twitter, while the Danish dataset is from Facebook, Reddit, and a local newspaper. The task also produced a weakly-supervised English dataset.

\end{itemize}

\section{Experiments}

\subsection{Experiment Settings}
We consider three training setups: (i) \textit{zero-shot} with no fine-tuning in the target, (ii) \textit{domain adaptation} in which we consider training on a set of auxiliary languages and test on a held-out language set, and (iii) \textit{full} fine-tuning using the training samples from all languages. Our experimental setup also requires a {\em base model}, acquired from a pre-trained model, which is fine-tuned on the source (English) language. 

For the cross-lingual analysis, we experiment on eight low-resource languages and one high-resource English. For the experiments on HASOC20 and HatEval19, we utilize the English samples in the training set to train the base model. For the experiments on SemEval2020 and HaSpeedDe20, we use Founta-EN as English training data is not available. 

The meta-training samples are retrieved from the source languages' \textit{validation sets} as well as the auxiliary and target languages' \textit{training sets}. The source languages' \textit{training sets} are used to train the {\em base model}. Across all the experiments, we aggregate samples from each language and then generate meta-tasks to be used in meta-training. We sample task triplets with a specified number of shots, $K = L$, where $K=32$.

We utilize two multilingual transformer-based encoders as base models: (i) $\bf mBERT$ \cite{devlin2019bert} (\texttt{bert-base-multilingual-uncased}), and (ii) $\bf XLM-R$ \cite{conneau-etal-2020-unsupervised} (\texttt{xlm-roberta-base}). For both models, the output from the {\textit{pooler}} layer is fed to a 2-layer FFN classification head.

\subsection{Baselines}
$\modelname$ is a model-agnostic algorithm for low-resource cross-lingual adaptation. Therefore, we explore the following baselines for our experiments: 
\begin{itemize}[leftmargin=*]
    \item \textbf{Fine-tuning}: Standard LM fine-tuning is adapted as a baseline for both zero-shot and few-shot scenarios. Here, we fine-tune on \textbf{mBERT} and \textbf{XLM-R}.
    \item \textbf{X-MAML} \cite{nooralahzadeh-etal-2020-zero}: Cross-lingual episodic meta-training using auxiliary language composition, including pairs.  We initialize the meta-learner from the base model obtained from English fine-tuning. The meta-learned model is then evaluated on the test set in the target language. Our implementation only considers one auxiliary language. 
\end{itemize}

\subsection{Variations of $\modelname$}
Besides benchmarking the baselines, we formulate and evaluate different variations of $\modelname$:

\begin{itemize}[leftmargin=*]
    \item {\bf $\modelname_{\textit{zero}}$:} Simulated meta-training on a batch of tasks created from aggregated samples in source language `validation set' and `training set' from auxiliary language. It provides zero-shot as training choice in Algorithm ~\ref{alg:HateMAML} (\S~ \ref{sec:algorithm}). This is similar to X-MAML\cite{nooralahzadeh-etal-2020-zero}, but it uses both source and auxiliary languages for meta-training, while X-MAML only uses auxiliary language samples.
    \item {\bf $\modelname_{\textit{self-training}}$:} Choose a target language for self-training on silver data. First, we generate silver labels and keep only 300 samples. Next, we apply $\modelname$ in an iterative manner to the silver-labeled data from the target language as explained in Section ~\ref{sec:selftrain}.
    \item  {\bf $\modelname_{ft}$:} We generate meta-task from a set of available languages and apply $\modelname$. This is very similar to fine-tuning aggregated training examples from the selected language set. 
\end{itemize}

\subsection{Implementation Details}
\label{app:implementation}
For implementing the transformer-base models, we use the HuggingFace \texttt{transformers} library\footnote{https://huggingface.co/transformers/}. The chosen pre-trained model is initialized from pre-trained weights provided in the \texttt{transformers} library. The classification heads are implemented using the pytorch\footnote{https://pytorch.org/} linear layers, initialized randomly. For implementing the meta-learning features (for example, first-order approximation), we use the \texttt{learn2learn} library\footnote{https://github.com/learnables/learn2learn}.

\begin{table*}[!ht]
    \caption{\textit{Zero-shot} evaluation on target low-resource languages in several multilingual hate speech datasets. Here $\clubsuit$ indicates that the pre-trained model is initialized with BERT, and $\spadesuit$ refers to XLM-R initialization. All the experiments (rows containing either $\clubsuit$ or $\spadesuit$ symbol) require a {\em base model} that is created using standard fine-tuning on a high-resource (EN) training set. Empty cells $(-)$ indicate that meta-learning experiments are impossible due to the unavailability of the required auxiliary language. In each cell ($x_y$), $x$ and $y$ denote the mean and standard deviation of macro-F1 values across five runs by varying the random seed initialization.}
    \label{tab:experiments_zeroshot}
    \adjustbox{max width=1\linewidth}{
    \begin{tabular}{p{3cm}cc|cccc|c|cc|c}
        \toprule
        \multirow{2}{*}{\bf Model} &  \multirow{2}{*}{\bf Aux lang} & \textbf{HatEval19} & \multicolumn{4}{c}{\bf SemEval20} & \multicolumn{1}{c}{\bf HaSpeedDe20} & \multicolumn{2}{c}{\bf HASOC20} &  \multirow{2}{*}{\bf AVG}\\
        \cmidrule{3-10}
        & & ES & AR & DA & EL & TR & $IT_{news}$ $IT_{tweets}$ & HI & DE \\

        \midrule
        
        mBERT &-& $0.499_{0.025}$ & $0.448_{0.000}$ & $0.650_{0.000}$ & $0.457_{0.000}$ & $0.452_{0.010}$ & $0.436_{0.023}$ $0.489_{0.25}$ & $0.426_{0.000}$ & $0.414_{0.000}$ & $0.473_{0.007}$ \\
        $\clubsuit X-MAML$ & \checkmark &-& $0.562_{0.024}$ & $0.654_{0.017}$ & $0.501_{0.040}$ & $0.585_{0.002}$ &-& $0.455_{0.015}$ & $0.629_{0.032}$ & $ 0.564_{0.021}$ \\
        $\clubsuit \modelname_{zero}$ & \checkmark&-& $0.583_{0.010}$ & $0.668_{0.017}$ & $0.532_{0.004}$ & $0.601_{0.005}$ &-& $0.448_{0.073}$ & $0.745_{0.046}$ & $\bf 0.592_{0.026}$\\ 
        $\clubsuit \modelname_{self\_training}$ &-& $\bf0.525_{0.033}$ & $0.529_{0.019}$ & $ 0.657_{0.046}$ & $0.547_{0.021}$ & $0.586_{0.002}$ &$0.569_{0.031}$ $\bf0.662_{0.056}$&  $0.509_{0.029}$ & $0.572_{0.048} $ & $0.562_{0.028}$\\
        
        \arrayrulecolor{black!30}\midrule

        XLM-R &-& $0.421_{0.064}$ & $0.524_{0.000}$ & $0.683_{0.000}$ & $0.529_{0.000}$ & $0.478_{0.000}$ & $0.457_{0.028},0.593_{0.039}$ & $0.448_{0.000}$ & $0.576_{0.000}$ & $0.523_{0.014}$\\
        $\spadesuit X-MAML$ & $\checkmark$ &-& $0.647_{0.007}$ & $0.699_{0.007}$ & $0.604_{0.023}$ & $0.527_{0.014}$ &-& $0.473_{0.010}$ & $0.672_{0.007}$ & $0.604_{0.011}$\\
        $\spadesuit \modelname_{zero}$ & $\checkmark$ &-& $0.651_{0.007}$ & $0.735_{0.006}$ & $0.619_{0.003}$ & $0.580_{0.028}$ &-& $0.497_{0.017}$ & $0.650_{0.013}$ & $\bf 0.622_{0.012}$ \\ 
        $\spadesuit \modelname_{self\_training}$ &-& $\bf0.436_{0.0846}$ & $0.621_{0.048}$ & $0.688_{0.036}$ & $0.635_{0.042}$ & $0.600_{0.016}$ &$0.492_{0.020}$ $0.637_{0.021}$& $0.598_{0.003}$ & $0.661_{0.005}$ & $0.591_{0.031}$\\
        \arrayrulecolor{black}\bottomrule[1.25pt]
    \end{tabular}
    }

\end{table*}

\begin{table*}[!ht]
    \caption{Domain Adaptation experiments on both fine-tuning and meta-training. We use languages = \{EN,ES,IT,DA,DE\} for training and evaluate on all eight languages. Here, first block means the pre-trained model is initialized with BERT and the second block refers to XLM-R initialization.}
    \label{tab:experiments_domain}
    \adjustbox{max width=1\linewidth}{
    \begin{tabular}{p{1.5cm}lc|cccc|c|cc|c}
        \toprule
        \multirow{2}{*}{\bf Model} &  \multirow{2}{*}{\bf data} & \textbf{HatEval19} & \multicolumn{4}{c}{\bf SemEval20} & \multicolumn{1}{c}{\bf HaSpeedDe20} & \multicolumn{2}{c}{\bf HASOC20} & \multirow{2}{*}{\bf{AVG}} \\
        \cmidrule{3-10}
        & & ES & AR & DA & EL & TR & $IT_{news}$, $IT_{tweets}$ & HI & DE \\
        \midrule

        $\clubsuit FT$ & 1024  & $0.684_{0.022}$ & $0.490_{0.035}$ & $0.711_{0.020}$ & $0.533_{0.012}$ & $0.504_{0.027}$ & $0.666_{0.035}, 0.703_{0.014}$ & $0.537_{0.023}$ & $0.735_{0.046}$ & $0.618_{0.098}$ \\
        $\clubsuit FT$& 2048 & $0.727_{0.049}$ & $0.517_{0.041}$ & $0.726_{0.018}$ & $0.515_{0.022}$ & $0.541_{0.024}$ & $0.667_{0.035}, 0.695_{0.031}$ & $0.464_{0.038}$ & $0.781_{0.010}$ & $0.626_{0.115}$ \\
        $\clubsuit FT$& 4096 & $0.753_{0.024}$ & $0.493_{0.037}$ & $0.753_{0.021}$ & $0.510_{0.035}$ & $0.546_{0.030}$ & $0.658_{0.041}, 0.736_{0.010}$ & $0.462_{0.041}$ & $0.805_{0.005}$ & $0.635_{0.130}$ \\
        $\clubsuit \modelname_{ft}$ & 1024 & $0.709_{0.006}$ & $0.495_{0.059}$ & $0.723_{0.022}$ & $0.545_{0.015}$ & $0.539_{0.032}$ & $0.607_{0.058}, 0.711_{0.004}$ & $0.498_{0.030}$ & $0.766_{0.013}$ & $0.622_{0.106}$ \\
        $\clubsuit \modelname_{ft}$ & 2048 & $0.737_{0.019}$ & $0.546_{0.009}$ & $0.724_{0.019}$ & $0.559_{0.017}$ & $0.556_{0.014}$ & $0.688_{0.031}, 0.722_{0.026}$ & $0.474_{0.021}$ & $0.795_{0.011}$ & $0.645_{0.108}$ \\
        $\clubsuit \modelname_{ft}$ & 4096 & $0.746_{0.018}$ & $0.556_{0.020}$ & $0.748_{0.025}$ & $0.559_{0.008}$ & $0.567_{0.009}$ & $0.695_{0.017}, 0.702_{0.035}$ & $0.475_{0.009}$ & $0.802_{0.015}$ & $\bf0.650_{0.109}$ \\
        \midrule
        
        $\spadesuit FT$ & 1024 & $0.521_{0.154}$ & $0.517_{0.088}$ & $0.574_{0.111}$ & $0.484_{0.047}$ & $0.517_{0.079}$ &  $0.521_{0.135}, 0.492_{0.164}$ & $0.492_{0.072}$ & $0.580_{0.162}$ & $0.522_{0.113}$ \\
        $\spadesuit FT$ & 2048 & $0.561_{0.180}$ & $0.482_{0.045}$ & $0.623_{0.149}$ & $0.507_{0.047}$ & $0.527_{0.077}$ & $0.540_{0.145}, 0.507_{0.188}$ & $0.438_{0.025}$ & $0.638_{0.207}$ & $0.536_{0.137}$ \\
        $\spadesuit FT$ & 4096 & $0.579_{0.191}$ & $0.468_{0.024}$ & $0.626_{0.146}$ & $0.513_{0.060}$ & $0.500_{0.060}$ & $0.570_{0.167}, 0.536_{0.182}$ & $0.431_{0.016}$ & $0.646_{0.211}$ & $0.541_{0.142}$ \\
        $\spadesuit \modelname_{ft}$ &  1024 & $0.725_{0.011}$ & $0.608_{0.014}$ & $0.716_{0.030}$ & $0.582_{0.027}$ & $0.624_{0.004}$ & $0.664_{0.042}, 0.732_{0.013}$ & $0.552_{0.018}$ & $0.769_{0.008}$ & $0.663_{0.075}$ \\
        $\spadesuit \modelname_{ft}$ &  2048 & $0.726_{0.022}$ & $0.637_{0.035}$ & $0.749_{0.019}$ & $0.609_{0.023}$ & $0.614_{0.029}$ & $0.663_{0.031}, 0.681_{0.046}$ & $0.510_{0.032}$ & $0.806_{0.010}$ & $\bf 0.666_{0.088}$ \\
        $\spadesuit \modelname_{ft}$ & 4096 & $0.738_{0.021}$ & $0.562_{0.061}$ & $0.745_{0.043}$ & $0.595_{0.019}$ & $0.586_{0.017}$ & $0.702_{0.009}, 0.700_{0.034}$ & $0.481_{0.025}$ & $0.805_{0.007}$ & $0.657_{0.104}$  \\
        \bottomrule
    \end{tabular}
    }
\end{table*}

\begin{table*}[!ht]
\caption{Fine-tuning and meta-training evaluation on multilingual hate speech datasets. We report F1 score. Here, first block means the pre-trained model is initialized with BERT and the second block refers to XLM-R initialization.}
    \label{tab:experiments_finetune}
    \adjustbox{max width=1\linewidth}{
    \begin{tabular}{p{1.5cm}lc|cccc|c|cc|c}
        \toprule
        \multirow{2}{*}{\bf Model} &  \multirow{2}{*}{\bf data} & \textbf{HatEval19} & \multicolumn{4}{c}{\bf SemEval20} & \multicolumn{1}{c}{\bf HaSpeedDe20} & \multicolumn{2}{c}{\bf HASOC20} & \multirow{2}{*}{\bf{AVG}} \\
        \cmidrule{3-10}
        & & ES & AR & DA & EL & TR & $IT_{news}$, $IT_{tweets}$ & HI & DE \\
        \midrule

        $\clubsuit FT$ & 1024 & $0.663_{0.009}$ & $0.721_{0.016}$ & $0.712_{0.027}$ & $0.700_{0.080}$ & $0.610_{0.053}$ & $0.629_{0.039}, 0.686_{0.009}$ & $0.590_{0.029}$ & $0.760_{0.017}$ & $0.675_{0.064}$ \\
         $\clubsuit FT$ & 2048 & $0.718_{0.024}$ & $0.765_{0.020}$ & $0.737_{0.028}$ & $0.717_{0.041}$ & $0.652_{0.050}$ & $0.652_{0.037}, 0.708_{0.021}$ & $0.609_{0.054}$ & $0.796_{0.012}$ & $0.706_{0.065}$\\
         $\clubsuit FT$ & 4096 & $0.717_{0.017}$ & $0.785_{0.024}$ & $0.747_{0.037}$ & $0.722_{0.007}$ & $0.670_{0.035}$ & $0.680_{0.045}, 0.684_{0.068}$ & $0.584_{0.071}$ & $0.803_{0.004}$ & $0.712_{0.077}$ \\
        $\clubsuit \modelname_{ft}$ & 1024 & $0.693_{0.015}$ & $0.726_{0.025}$ & $0.692_{0.018}$ & $0.685_{0.044}$ & $0.626_{0.018}$ & $0.593_{0.026}, 0.702_{0.020}$ & $0.583_{0.020}$ & $0.743_{0.023}$ & $0.671_{0.059}$ \\
        $\clubsuit \modelname_{ft}$ & 2048 & $0.735_{0.013}$ & $0.779_{0.011}$ & $0.716_{0.013}$ & $0.732_{0.026}$ & $0.665_{0.007}$ & $0.670_{0.019}, 0.730_{0.011}$ & $0.627_{0.006}$ & $0.789_{0.013}$ & $0.716_{0.052}$ \\
        $\clubsuit \modelname_{ft}$ & 4096 & $0.757_{0.018}$ & $0.816_{0.012}$ & $0.752_{0.010}$ & $0.742_{0.022}$ & $0.689_{0.012}$ & $0.643_{0.042}, 0.741_{0.010}$ & $0.627_{0.016}$ & $0.793_{0.009}$ & $\bf 0.729_{0.063}$ \\
        \midrule
        
        $\spadesuit FT$ & 1024 & $0.630_{0.063}$ & $0.680_{0.106}$ & $0.774_{0.028}$ & $0.595_{0.064}$ & $0.593_{0.077}$ & $0.605_{0.081}, 0.643_{0.029}$  & $0.621_{0.036}$  & $0.735_{0.059}$ & $0.653_{0.085}$ \\
        $\spadesuit FT$ & 2048 & $0.684_{0.019}$ & $0.730_{0.043}$ & $0.782_{0.022}$ & $0.673_{0.061}$ & $0.671_{0.032}$ & $0.683_{0.039}, 0.629_{0.058}$ & $0.641_{0.043}$ & $0.800_{0.006}$ & $0.699_{0.067}$ \\
        $\spadesuit FT$ & 4096 & $0.704_{0.015}$  & $0.806_{0.021}$ & $0.774_{0.008}$ & $0.705_{0.021}$ & $0.684_{0.013}$ & $0.651_{0.033}, 0.679_{0.023}$ & $0.632_{0.022}$ & $0.806_{0.011}$ & $0.716_{0.064}$ \\
        $\spadesuit \modelname_{ft}$ & 1024 & $0.728_{0.004}$ & $0.781_{0.016}$ & $0.736_{0.001}$ & $0.703_{0.004}$ & $0.695_{0.001}$ & $0.608_{0.015}, 0.741_{0.018}$ & $0.664_{0.010}$ & $0.789_{0.000}$ & $0.716_{0.056}$ \\
        $\spadesuit \modelname_{ft}$ & 2048 & $0.742_{0.012}$ & $0.803_{0.002}$ & $0.733_{0.003}$ & $0.692_{0.018}$ & $0.695_{0.024}$ & $0.652_{0.015}, 0.748_{0.007}$ & $0.672_{0.005}$ & $0.823_{0.009}$ & $0.729_{0.057}$ \\
        $\spadesuit \modelname_{ft}$ & 4096 & $0.755_{0.005}$ & $0.832_{0.010}$ & $0.736_{0.022}$ & $0.711_{0.019}$ & $0.706_{0.033}$ & $0.694_{0.036}, 0.738_{0.022}$ & $0.657_{0.016}$ & $0.812_{0.008}$ & $\bf 0.738_{0.056}$ \\
        \bottomrule
    \end{tabular}
    }
\end{table*}

%
%

\section{Results and analysis}
\label{sec:results and analysis}

\subsection{Zero-shot Experiments}
\label{sec:zero-shot-analysis}
Experiments in the zero-shot setting can be found in Table \ref{tab:experiments_zeroshot}. The mBERT initialization uses the $\clubsuit$ sign, and $\spadesuit$ refers to XLM-R initialization.
The reported results of all the experiments are average values across five random runs. We report the average and standard deviation of the \textbf{F1-score (macro)}. $\modelname_{zero}$ outperforms baseline models including X-MAML with both initializations ($\clubsuit$ and $\spadesuit$). We achieve the best overall score with an average of 11\% over mBERT, 7\% over XLM-R, 7\% over $\clubsuit$ X-MAML. Baseline models, mBERT and XLM-R, have poor \textit{zero-shot} performance. The use of auxiliary language to obtain the best meta-learner produces rewarding outcomes for all the experiments. Even though X-MAML uses a similar cross-lingual, auxiliary language-based, meta-training setting to ours, $\modelname$ gives superior performance, likely due to the domain-adaptive training in \modelname. To summarize, in most cases, $\modelname$'s auxiliary to target transfer improves the results of zero-shot learning substantially compared to the standard fine-tuning baselines.

We observe that the XLM-R-initialized models perform better than mBERT in zero-shot prediction. Even the best performing $\clubsuit \modelname$ model has a small gap between XLM-R zero-shot performance. One explanation is that XLM-R is trained on a relatively large corpus and a deeper transformer architecture, giving it a strong pre-training benefit for zero-shot transfer learning. We observe similar trends in performance with $\modelname$ when initialized with XLM-R base models across all the datasets. For the SemEval20 Danish (DA) test set, it is noted that the XLM-R model performs the best compared to other meta-learning models. One reason could be that this dataset was created from Facebook, Reddit, and news sources, unlike the other datasets that have been created from Twitter. 

{\bf Language similarity and cross-lingual transfer.}
We carefully analyze the support that each target language receives from different auxiliary languages. The SemEval20 dataset, with the largest number of possible (auxiliary, target) language pairs, shows the benefit of meta-training on auxiliary language for both X-MAML and $\modelname_{zero}$ models. Meta-training helps in increasing the zero-shot performance for SemEval20 from 0.473 avg. F1 (mBERT) to 0.592 F1 ($\clubsuit \modelname$). X-MAML also shows substantial improvement for both initializations. 
We notice that SemEval20 Turkish (TR) tends to be the best auxiliary language dataset in almost all cases for both base models (see  Table ~\ref{tab:aux_langs}).
For HASOC20 experiments, we notice small gain in F1-score while training on auxiliary samples. The languages -- German (DE) and Hindi (HI), do not seem to be a good mix, coming from two distant language families (details of family languages is provided in Table~\ref{tab:langs_iso}).

\begin{table}[!ht]
    \caption{List of languages and their ISO codes used in our experiments.}
    \centering
    \begin{tabular}{ccc}
    \toprule
        \textbf{Language} & \textbf{ISO 639-1 code} & \textbf{Family} \\
        \midrule
        Arabic & AR & Afro-Asiatic \\
        \midrule
        Danish & DA &  IE: Germanic \\
        \midrule
        German & DE & IE: Germanic \\
        \midrule
        Greek & EL & IE: Greek \\
        \midrule
        English & EN &  IE: Germanic \\
        \midrule
        Hindi & HI &  IE: Indo-Aryan \\
        \midrule
        Spanish & ES & IE: Italic  \\
        \midrule
        Turkish & TR & Turkic \\
    \bottomrule
    \end{tabular}
    \label{tab:langs_iso}
\end{table}

\begin{table}[!t]
    \caption{Zero-shot experiments and results using auxiliary languages on SemEval2020 dataset. All the results are produced from XLM-R initialization and average over 5 runs.}
    \centering
    \adjustbox{max width=1\linewidth}{
    \begin{tabular}{cccccc}
        \midrule
        \multirow{2}{*}{Model} & \multirow{2}{*}{Auxiliary lang} & \multicolumn{4}{c}{Target lang}  \\
        \cmidrule{3-6}
        & & AR & TR & DA & EL \\
        \midrule
        $\modelname_{zero}$ & AR & - & $0.553_{0.026}$ & $0.718_{0.011}$ & $0.591_{0.008}$\\
        $\modelname_{zero}$ & TR & $0.651_{0.007}$ & - & $0.738_{0.012}$ & $0.619_{0.003}$\\
        $\modelname_{zero}$ & DA & $0.628_{0.022}$ & $0.580_{0.028}$ & - & $0.553_{0.026}$\\
        $\modelname_{zero}$ & EL & $0.530_{0.017}$  & $0.494_{0.021}$ & $0.657_{0.044}$ & -  \\
        \bottomrule
    \end{tabular}
    }
    \label{tab:aux_langs}
\end{table}

{\bf Self-training improves zero-shot performance.}
\label{sec:self-training-analysis}
For datasets that have no auxiliary language, i.e., HatEval19 and HaSpeedDe20, self-training $\modelname$ is a convenient algorithm to maximize the performance. It mitigates the need for auxiliary language or a labeled target language training set. We find that $\modelname_{self-training}$ produces comparable performance to meta-training in auxiliary $\modelname_{zero}$ (see Table~\ref{tab:experiments_zeroshot}). The gain is achieved from meta-training on silver labels of the target using multiple iterations of self-refinement. We also utilize source language validation set while self-training ($\modelname_{\textit{self-training}}$). The intuition is to have some gold-labeled samples to balance between noise and ground truth. We observe a slight improvement given the source ground-truth labels in training. 


\subsection{Domain Adaptation Experiments}
\label{sec:domain-analysis}
Table ~\ref{tab:experiments_domain}  reports the results of domain adaptation experiments. We first select two language families for training: Germanic and Romance. The languages in the selected languages families, namely EN, DA, DE, ES, and IT, are considered auxiliary and used in the training set. The trained model is evaluated on the held-out languages. We use training samples in increasing order to evaluate the model performance in the order of available data. Each language's varying training data size has three values, i.e., $1024$, $2048$, and $4096$. However, if a language does not contain the selected number of samples, i.e., DE has only $2373$ training samples, we capped it to the maximum of found training samples. It can be seen that $\modelname$ produces superior results on average compared to fine-tuning. We find that domain-adaptive meta-training in $\modelname_{ft}$ ($\clubsuit$ or $\spadesuit$) has a consistently better F1 score than fine-tuning. We achieve an overall improvement of $3\%$ over mBERT, $24\%$ over XLM-R while training $2048$ samples per language. Both baseline models face a performance drop in held-out languages, in which XLM-R faces a significant performance drop. To summarize, standard fine-tuning does not generalize well during training on the held-out target languages. On the other hand, $\modelname$'s domain-adaptive training helps retain consistent performance on both auxiliary and target languages.

\subsection{Full Fine-tuning vs Meta-training}
To make one model support all languages, we evaluate a meta-training strategy while training on data available in eight languages. We train a model varying the training data size and evaluate performance under increasing data availability. We create a set of meta-tasks by gathering support and query sets from eight languages. We also fine-tune aggregated training data. In Table \ref{tab:experiments_finetune}, we show how $\modelname_{ft}$ performs in comparison to mBERT and XLM-R fine-tuning. In general, we notice an upward trend while increasing the number of training data for all languages. As expected, the meta-learned initialized $\modelname_{ft}$ models perform better than the standard fine-tuning. $\modelname_{ft}$ outperforms on average for increasing the order of available data. We observe that $\modelname_{ft}$ gains in F1 score for all languages -- 2.4\% and 3\% improvement over baselines mBERT and XLM-R using $4096$ samples. To improve performance further, we can increase training data, if available. This suggests that meta-training can be an alternative to standard fine-tuning for hate speech detection.


\section{Conclusion}

Our extensive experiments show that $\modelname$ is able to perform well in both zero-shot and few-shot hate speech detection by improving cross-lingual transfer in target languages. $\modelname$ can be trained with or without labeled data. This model-agnostic approach supports multilingual encoders with shared embeddings such as mBERT and XLM-R as base learners. The proposed zero-shot self-refining technique adapts a base learner to be an effective predictor in the target, reducing the need for ground-truth labels in fine-tuning. To further improve cross-lingual transfer, we use a few labeled samples in the target language (few-shot training), which helps to adapt fast and boosts fine-tuning performance. Additionally, our method benefits from task-specific learner evaluation on a (virtual) domain test (query) set at each episode that mimics target domain evaluation during cross-lingual meta-training. The intuition of the domain query set is to achieve training generalization in unseen target languages and domains. We also found that meta-learner adaptation effectively supports all available languages using a single predictor model, making it highly suitable for detecting hate speech in multiple languages and domains.

In this paper, our major contributions are two-fold. (i) We proposed $\modelname$, a novel model-agnostic cross-lingual meta-training algorithm for hate speech detection that supports resource maximization. We evaluated $\modelname$ on five benchmark datasets across eight languages and varying domains. We showed that cross-lingual meta-training outperforms state-of-the-art fine-tuning baselines in both zero-shot and fine-tuning settings. We also found that meta-adaptation is effective in cross-domain evaluation, making it highly suitable for detecting cross-domain hate speech. (ii) We devised a self-training procedure that aids in hate speech detection in extreme data-scarce scenarios. In future, we aim to explore multi-class hate speech detection and meta-training on a set of heterogeneous tasks such as aggression, target identification, code-mixing, etc. We hope our work will motivate modeling fast adaptation for cross-lingual training and zero-shot hate speech detection.

\bibliographystyle{IEEEtran}
\bibliography{ref}

\begin{IEEEbiography}[{\includegraphics[width=1in,height=1.25in,clip,keepaspectratio]{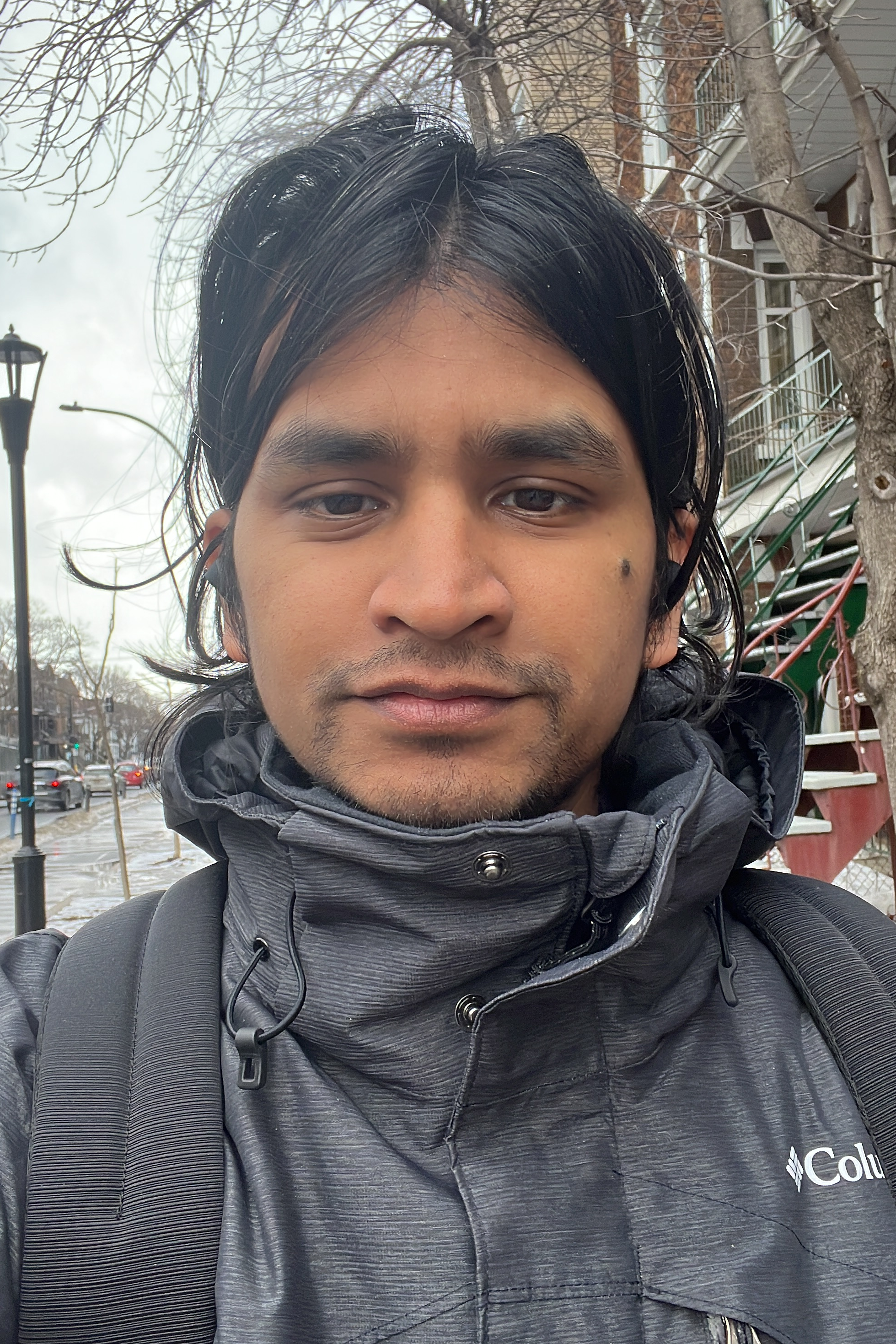}}]{Rabiul Awal} earned a Bachelor's degree in Computer Science from Noakhali Science and Technology University and a Master's degree from the University of Saskatchewan. His research focuses on the intersection of social NLP, deep learning, and language-vision. Currently, he is a research intern at Mila, Montreal.
\end{IEEEbiography}
\vskip -2\baselineskip plus -1fil
\begin{IEEEbiography}[{\includegraphics[width=1in,height=1.25in,keepaspectratio]{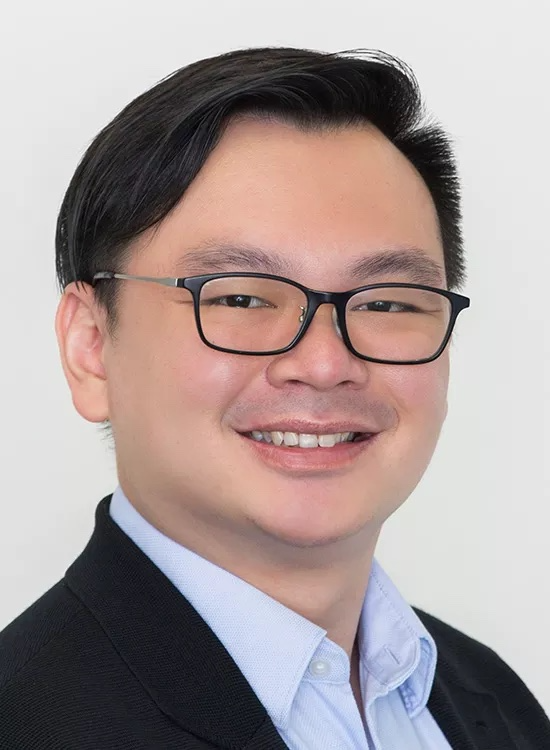}}]{Roy Ka-Wei Lee} is an Assistant Professor at the Information Systems Technology and Design Pillar, Singapore University of Technology and Design. He is a faculty of the transformative Design and Artificial Intelligence programme. His research lies in the intersection of data mining, machine learning, social computing, and natural language processing. He is leading the Social AI Studio, a research group that focuses on designing next-generation social artificial intelligence systems. He has published in top-tier venues in data mining and computation linguistics domains. He serves on the program committees of multiple top data mining and natural language processing conferences.
\end{IEEEbiography}
\vskip -2\baselineskip plus -1fil
\begin{IEEEbiography}[{\includegraphics[width=1in,height=1.25in,clip,keepaspectratio]{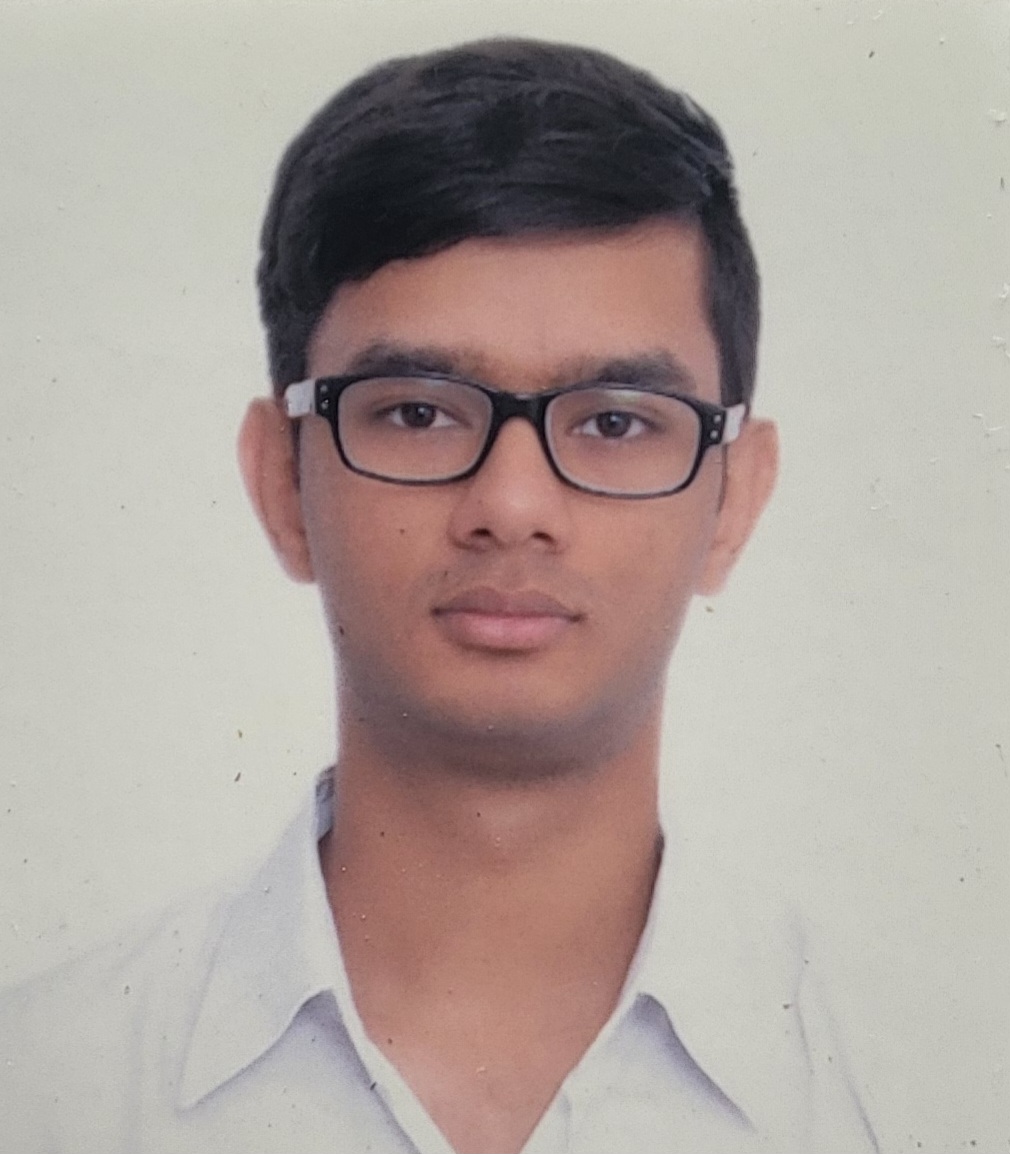}}]{Eshaan Tanwar}
Eshaan Tanwar received his bachelor of technology from Delhi Technological University, New Delhi, India.
He is currently a Research Associate in the Laboratory for Computational Social Systems (LCS2). His research interests include Social Computing, Natural Language Processing, and Machine Learning.
\end{IEEEbiography}
\vskip -2\baselineskip plus -1fil
\begin{IEEEbiography}[{\includegraphics[width=1in,height=1.25in,clip,keepaspectratio]{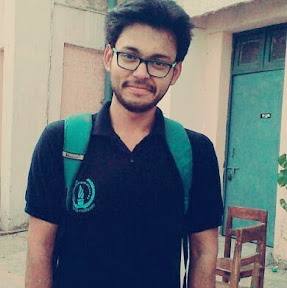}}]{Tanmay Garg}
Tanmay Garg completed his Master's in Technology from Indiraprastha Institute of Information Technology, Delhi in 2019. He is currently working as a Research Engineer. His research interests include Hate Speech, Bias in Hate Speech and Natural Language Processing.
\end{IEEEbiography}
\vskip -2\baselineskip plus -1fil
\begin{IEEEbiography}[{\includegraphics[width=1in,height=1.25in,clip,keepaspectratio]{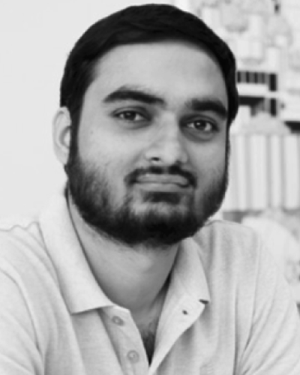}}]{Tanmoy Chakraborty} is an Associate Professor and a Ramanujan Fellow in the Department of Electrical Engineering at the Indian Institute of Technology Delhi (IIT Delhi), India since September 2022. Prior to joining IIT Delhi, he held positions as an Assistant Professor (May 2017 - Dec 2021) and an Associate Professor (Jan 2022 - Aug 2022) in the Department of CSE at IIIT Delhi, India. He also served as the head of the Infosys Centre for AI (CAI) at IIIT Delhi. Tanmoy Chakraborty earned his PhD as a Google PhD scholar from the Department of CSE at IIT Kharagpur, India in September 2015. Following this, he worked as a Postdoctoral Researcher at the University of Maryland, College Park. He leads LCS2, a research group that broadly works in  Natural Language Processing, Graph Neural Networks, and Social Computing. He is a senior IEEE member. 
\end{IEEEbiography}

\end{document}